%% file: emnlp2020.tex
\documentclass[11pt,a4paper]{article}
\usepackage[hyperref]{emnlp2020}
\usepackage{times}
\usepackage{latexsym}

\usepackage{microtype}

\aclfinalcopy 
\def\aclpaperid{3453} 

\usepackage[utf8]{inputenc}
\usepackage{caption}
\usepackage{graphicx}
\usepackage{latexsym}
\usepackage{booktabs}
\usepackage{multirow}
\usepackage{epstopdf}
\usepackage{balance}
\usepackage{subfigure}
\usepackage[noend,ruled]{algorithm2e}
\usepackage{microtype}
\usepackage{enumitem}
\usepackage{wrapfig,lipsum}
\usepackage{amsfonts}
\usepackage{amsmath}
\usepackage{float}
\restylefloat{table}

\usepackage{cleveref}

\newcommand{\dr}[1]{\textcolor{red}{[DR: #1]}}
\newcommand{\drc}[1]{\textcolor{red}{#1}}
\newcommand{\Xingyu}[1]{\textcolor{purple}{#1}}

\newcommand{\ignore}[1]{}

\DeclareMathOperator*{\argmax}{arg\,max}

\DeclareMathOperator{\mbert}{M-BERT}
\newcommand{\modelname}[0]{\texttt{QuEL}}
\newcommand{\chentsemodel}[0]{\texttt{xlwikifier}}
\newcommand{\shyammodel}[0]{\texttt{xelms}}
\newcommand{\hengmodel}[0]{\texttt{ELISA}}
\newcommand{\neubmodel}[0]{\texttt{PBEL\_PLUS}}

\newcommand{\ptm}[0]{\texttt{p(e|m)}}
\newcommand{\trans}[0]{\texttt{name\_trans}}
\newcommand{\pivot}[0]{\texttt{pivoting}}
\newcommand{\tsl}[0]{\texttt{translit}}

\newcommand{\query}[0]{\texttt{QuEL\_CG}}

\title{Design Challenges in
Low-resource Cross-lingual Entity Linking}
\ignore{
Design \drc{and Resource} Challenges \drc{in }
Low-resource Cross-lingual Entity Linking\\
\Xingyu{Design and Resource Challenges in Low-resource Cross-lingual Entity Linking: A Thorough Analysis?}\\
\dr{Down side of this suggestion: two lines; thoughts?}
\Xingyu{maybe without ''Resource"?} \dr{I added ``resource" with the hope to focus on this aspect, but we can live without it}}

\author{Xingyu Fu$^{1}$\thanks{\indent Both authors contributed equally to this work.}, Weijia Shi$^{2*}$, Xiaodong Yu$^{1}$, Zian Zhao$^{3}$, Dan Roth$^{1}$ \\
  $^1$University of Pennsylvania \\
  $^2$University of California Los Angeles \\
  $^3$Columbia University\\
  \texttt{\{xingyuf2, xdyu, danroth\}@seas.upenn.edu}\\ \texttt{swj0419@g.ucla.edu, zz2562@columbia.edu} \\}
\date{}

\begin{document}
\maketitle

\input{sections/0-abs}
\input{sections/1-intro}
\input{sections/2-related}
\input{sections/3-existing-limit}
\input{sections/4-method}
\input{sections/5-exp}
\input{sections/6-analysis}
\input{sections/7-conclud}
\input{sections/7.5-ack}
\bibliographystyle{acl_natbib}
\bibliography{emnlp2020,ccg.bib}

\appendix
\clearpage
\input{sections/8-appendix}

\end{document}

%% file: sections/0-abs.tex
\begin{abstract}
Cross-lingual Entity Linking (XEL), the problem of grounding mentions of entities in a foreign language text into an English knowledge base such as Wikipedia, has seen a lot of research in recent years, with a range of promising techniques. However,
current techniques do not rise to the challenges introduced by text in low-resource languages (LRL) and, surprisingly, fail to generalize to text not taken from Wikipedia, on which they are usually trained. \\
%
This paper provides a thorough analysis of low-resource XEL techniques, focusing on the key step of identifying candidate English Wikipedia
titles 
that correspond to a given foreign language mention. Our analysis indicates that current methods are limited by their reliance on Wikipedia's
interlanguage
links and thus suffer when the foreign language's Wikipedia is small.
We conclude that the LRL setting requires the use of outside-Wikipedia cross-lingual resources and present a simple yet effective zero-shot XEL system, \modelname, that utilizes search engines query logs. 
With experiments on 25 languages, 
\modelname~shows an average increase of \textbf{25\%} in gold candidate recall and of \textbf{13\%} in end-to-end linking accuracy over state-of-the-art baselines.\footnote{
Code
is available at: \url{http://cogcomp.org/page/publication_view/911}. The LORELEI data will be available through LDC.} 
\end{abstract}

%% file: sections/1-intro.tex
\section{Introduction}
\begin{figure}[t]
\centering
\includegraphics[width=1\linewidth]{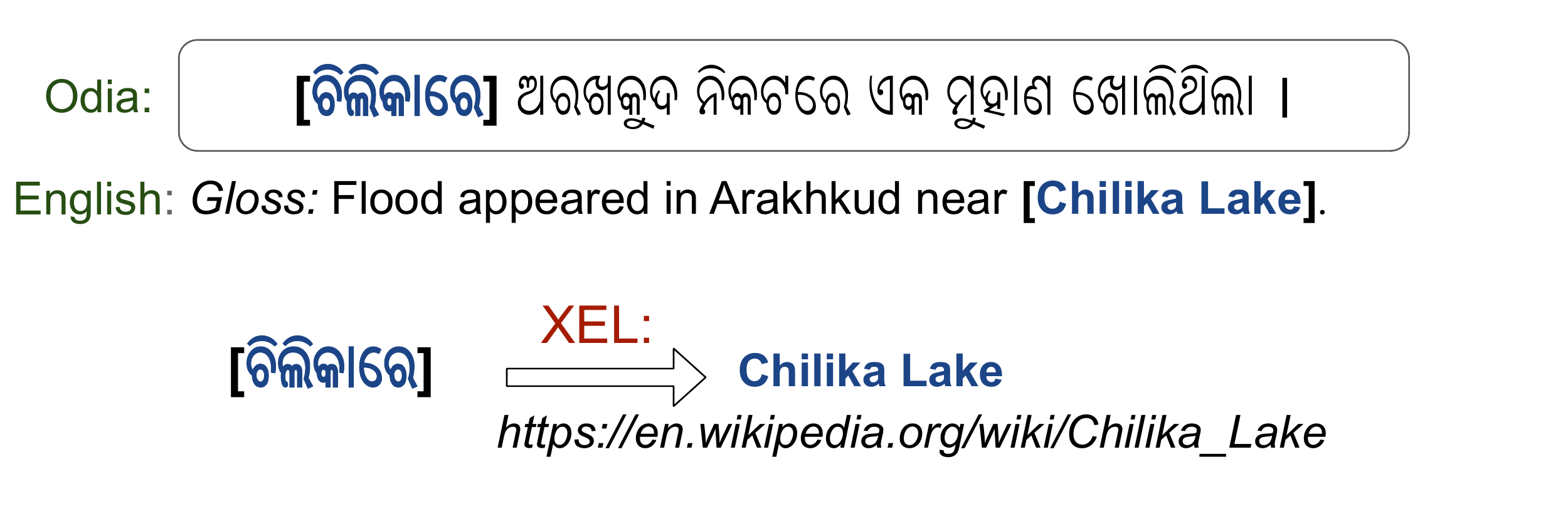}
\vspace{-0.4cm}
\vspace{-0.2cm}
\caption{The EXL task: in the given sentence we link ``Chilika Lake" to its corresponding English Wikipedia
}
\label{fig:xel_example}
\end{figure}

Cross-lingual Entity Linking (XEL) aims at grounding mentions written in a foreign (source) language ($SL$) into entries in a (target) language Knowledge Base (KB), which we consider here as the English Wikipedia following ~\newcite{pan2017cross,UpGuRo18,zhou2020improving}.
In Figure~\ref{fig:xel_example}, for instance, an Odia (an Indo-Aryan language in India) mention (``Chilika Lake") is linked to the corresponding English Wikipedia entry.
The XEL task typically involves two main steps: (1) candidate generation, retrieving a list of candidate KB entries for the mention, and (2) candidate ranking, selecting the most likely entry from the candidates.

While XEL techniques have been studied heavily in recent years, many challenges remain in the LRL setting.
Specifically, existing candidate generation methods perform well on Wikipedia-based dataset but fail to generalize beyond Wikipedia, to news and social media text. 
Error analysis on existing LRL XEL systems shows that the key obstacle is candidate generation. For example, 79.3\%-89.1\% of XEL errors in Odia can be attributed to the limitations of candidate generation.

\begin{figure}[h]
\centering
\includegraphics[width=1\linewidth]{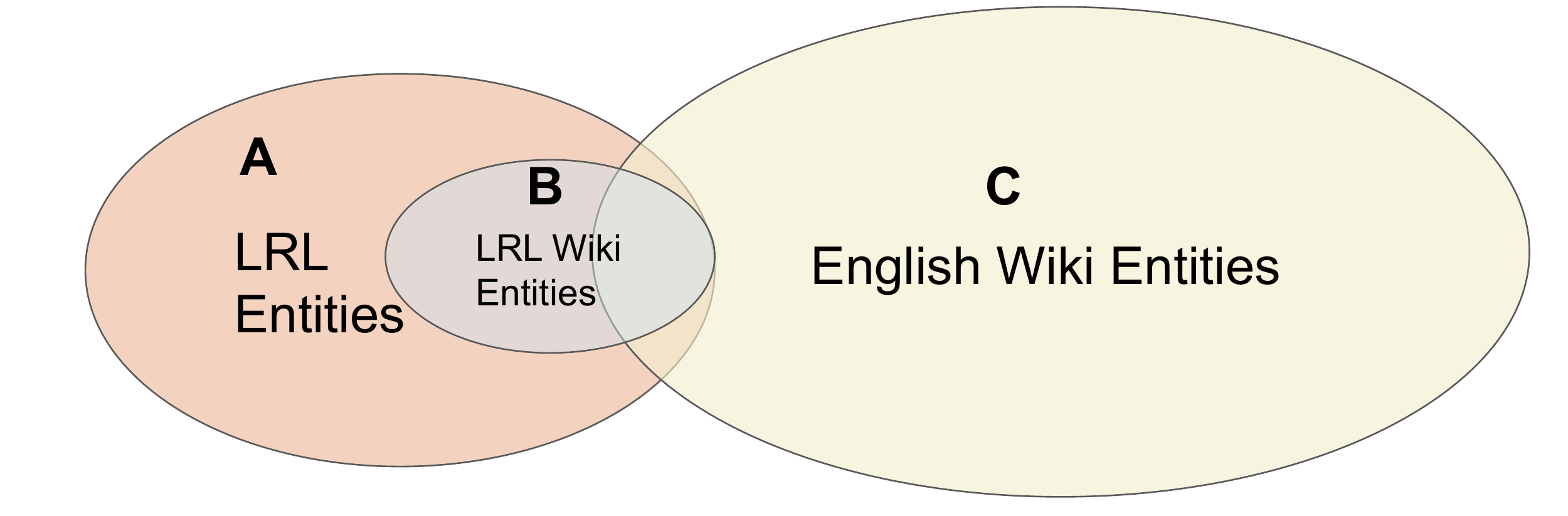}
\vspace{-0.4cm}
\vspace{-0.2cm}
\caption{
An illustration of Cross-lingual resources serving current XEL systems.
}
\label{fig:lingual_resource}
\end{figure}

In this paper, we present a thorough analysis of the limitations of several leading candidate generation methods. Although these methods adopt different techniques, we find that they all 
heavily rely on Wikipedia interlanguage links\footnote{\url{https://en.wikipedia.org/wiki/Help:Interlanguage_links}} as their cross-lingual resources. However, small $SL$ Wikipedia size limits their performance in the LRL setting.
As shown in Figure~\ref{fig:lingual_resource}, while the core challenge of LRL XEL is to link LRL entities ($A$) to candidates in the English Wikipedia ($C$), interlanguage links
only map a small subset of the LRL entities that appear in both LRL Wikipedia ($B$) and English Wikipedia.  
Therefore, methods that only leverage interlanguage links ($B\cap C$) as the main source of supervision cannot cover a wide range of entities. 
For example, the Amharic Wikipedia has 14,854 entries, but only 8,176 of them have interlanguage links to English.

Furthermore, as we show, existing candidate generation methods perform well on Wikipedia-based datasets but fail to generalize to outside-Wikipedia text such as news or social media. 


Our observations lead to the conclusion that the LRL setting necessitates the use of outside-Wikipedia cross-lingual resources. Specifically, we propose various ways to utilize the abundant query log from online search engines 
to compensate for the lack of supervision.
Similar to Wikipedia, a free online encyclopedia created by Internet users, 
Query logs (QL) provide
a free resource, collaboratively generated by a large number of users, and mildly curated.
However, it is orders of magnitude larger than Wikipedia.\footnote{ \url{https://www.internetlivestats.com/google-search-statistics/}} In particular, it includes all of Wikipedia cross-lingual resources as a subset since a search of an $SL$ mention leads to the English Wikipedia entity if the corresponding Wikipedia entries are interlanguage-linked.

The main part of this paper, Sec.~\ref{sec:limit}, presents a thorough \emph{method-wise}
evaluation and analysis of leading candidate generation methods, and quantifies their limitations as a function of $SL$ Wikipedia size and the size of the interlanguage cross-lingual resources.
Driven by these limitations, we propose and analyze (Sec.~\ref{sec:method}) \query, an improved candidate generation method utilizing QL, showing its advantages over the 
Wikipedia resources 
on LRL.
To exhibit a \emph{system-wise} XEL comparison, in Sec.~\ref{sec:exp}, we suggest
a simple yet efficient zero-shot XEL framework \modelname, that incorporates \query. \modelname~achieves an average of \textbf{25\%} increase in gold candidate recall, and \textbf{13\%} increase in end-to-end linking accuracy on outside-wikipedia text.

\ignore{
After a survey of relevant related work in Sec.~\ref{sec:related}, the main part of the paper, Sec.~\ref{sec:limit} presents a thorough \emph{method wise} evaluation and analysis of leading candidate generation methods of several SOTA systems, and quantifies their limitations as a function of $SL$ Wikipedia size and the size of the interlanguage cross-lingual resources. 
We also analyze \query, an improved candidate generation method utilizing QL proposed in Sec.~\ref{sec:method}, showing that it improves the recall of existing methods by 25.7\%. 

Building on this analysis, Sec.~\ref{sec:exp} suggests a simple yet efficient zero-shot XEL framework \modelname, that incorporates our new method \query, and exhibits a \emph{system wise} XEL comparison.
It is shown that \modelname~achieves an average of \textbf{25\%} increase in gold candidate recall, and an average of \textbf{13\%} increase in end-to-end linking accuracy on outside-wikipedia text.
We conclude with lessons learned and some future work.
}

%% file: sections/2-related.tex
\section{Related Work}
\label{sec:related}

\subsection{Background: Wikipedia resources}

\noindent    \textbf{Wikipedia title mappings}: 
    The mapping comes from Wikipedia interlanguage links\footnote{\url{https://en.wikipedia.org/wiki/Help:Interlanguage_links}} between the $SL$ and English, and uses Wikipedia articles titles as mappings.
    It directly links an $SL$ entity to an English Wikipedia entry without ambiguity. 
    
\noindent    \textbf{Wikipedia anchor text mappings}: A clickable text mention in Wikipedia articles is annotated with anchor text linking it to a Wikipedia entry.
    The following retrieving order: $SL$ anchor text $\rightarrow$
    $SL$ Wikipedia entry $\rightarrow$ English Wikipedia entry (where the last step is done via the Wikipedia interlanguage links), 
    allows one to build a bilingual title mapping from a $SL$ mentions 
    to Wikipedia English entries, resulting in a probabilistic mapping 
    with scores calculated using total counts~\cite{TsaiRo16b}.

\subsection{XEL systems for Low-resource languages}
We briefly survey key approaches to XEL below.

\noindent\textbf{Direct Mapping Based Systems}, including \chentsemodel~\cite{TsaiRo16b} and \shyammodel~\cite{UpGuRo18}, focus on building a $SL$ to English mapping to generate candidates. 
For candidate ranking, both \chentsemodel~ and \shyammodel~ combine supervision from multiple languages to learn a ranking model.

\noindent\textbf{Word Translation Based Systems} including \newcite{pan2017cross} and \hengmodel~\cite{zhang2018elisa}, extract $SL$ -- English name translation pairs and apply an unsupervised collective inference approach to link the translated mention.

\noindent\textbf{Transliteration Based Systems} include \newcite{TsaiRo18} and \tsl~\cite{UpKoRo18}.
\tsl~uses a sequence-to-sequence model and bootstrapping to deal with limited data. It is useful when the English and $SL$ word pairs have similar pronunciation.

\noindent\textbf{Pivoting Based Systems} including~\cite{rijhwani2019zero, zhou2019towards} and \neubmodel~\cite{zhou2020improving}, remove the reliance on $SL$ resources and use a \emph{pivot} language for candidate generation.
Specifically, they train the XEL model on a selected high-resource language, and apply it to $SL$ mentions through language conversion.

\subsection{Query Logs}
Query logs have long been used in many tasks such as across-domain generalization \cite{rud2011piggyback} for NER, and ontological knowledge acquisition\cite{alfonseca2010acquisition,pantel2012mining}.
In the English entity linking task, \newcite{6823700} pointed out that google query logs can be an efficient way to identify candidates. \newcite{Dredze:2010:EDK:1873781.1873813, Monahan2011CrossLingualCC} use search result as one of their methods for candidate generation on high resource language entity linking task. While earlier work indicates that query logs provide abundant information that can be used in NLP tasks, as far as we know, it has never been used in cross-lingual tasks as we do here. And, only search information has been used, while we suggest using Maps information too.

%% file: sections/3-existing-limit.tex
\section{Candidate Generation Analysis}
\label{sec:limit}
In this section, we analyze four leading candidate generation methods: \ptm, \chentsemodel, \trans, \pivot, and \tsl (see Table~\ref{tbl:method2baseline} for the XEL systems that use one or multiple of the methods) and discuss their limitations.
\input{tables/method_system_table}

\begin{figure*}[t]
\centering
\includegraphics[width=1\linewidth]{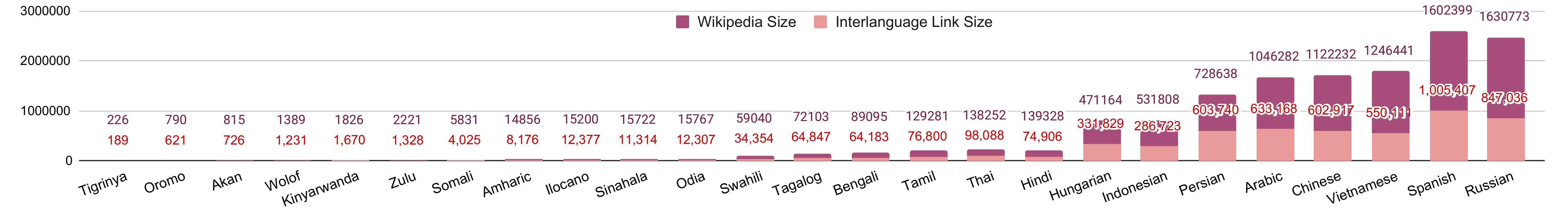}
\vspace{-0.6cm}
\caption{$SL$ Wikipedia size (number of articles) and Interlanguage links size (between $SL$ and English). We define low-resource languages (LRL) as Tigrinya to Odia, and high-resource ones (HRL) as Swahili to Russian.}
\label{fig:wiki_size_lor}
\end{figure*}

\vspace{-0.3cm}
\subsection{Candidate Generation Methods}
Each method listed in Table~\ref{tbl:method2baseline}, is discussed below along with the level of resources it requires.

\noindent        \textbf{\ptm}~\cite{TsaiRo16b} creates a direct probabilistic mapping table using Wikipedia title mappings and the anchor text mappings, between $SL$ and English.
        E.g., if an Oromo mention ``Itoophiyaatti” is the anchor text linked to an Oromo Wikipedia entity ``Itoophiyaa”\footnote{\url{https://om.wikipedia.org/wiki/Itoophiyaa}}, and ``Itoophiyaa” has an interlanguage link to English Wikipedia entity ``Ethiopia”, then ``Ethiopia” will be added as a candidate for the mention ``Itoophiyaatti”. 
        Thus, \ptm~follows a linking flow: $SL$ mention $\rightarrow$ $SL$ Wikipedia entity $\rightarrow$ English Wikipedia entity.23
         
\noindent        \textbf{\trans} (Name Translation) as introduced in \cite{pan2017cross, zhang2018elisa} performs word alignment on Wikipedia title mappings, to induce a fixed word-to-word translation table between $SL$ and English.
        For instance, to link the Suomi name ``Pekingin tekninen instituutti'' (Beijing Institute of Technology), it translates each word in the mention: (``Pekingin'' -- Beijing, ``tekninen'' -- Technology, ``instituutti'' -- Institute).
        At test time, after mapping each word in the given mention to English, it links the translation to English Wikipedia using an unsupervised collective inference approach.
        
\noindent        \textbf{\tsl}~\cite{UpKoRo18} trains a seq2seq model on
Wikipedia title mappings, to generate the English entity directly.
        
\noindent        \textbf{\pivot} to a related high-resource language (HRL)~\cite{zhou2020improving} attempts to generalize to unseen LRL mentions through grapheme or phoneme similarity between $SL$ and a related HRL.
        Specifically, it first finds a related HRL for the $SL$,
        and learns a XEL model on the related HRL,
        using Wikipedia title mappings and anchor text mappings.
        Then it applies the model on $SL$ mentions.
        If $SL$ and related HRL share a script, showing grapheme similarity, it treats $SL$ mentions as related HRL mentions at test time.
        Otherwise, if $SL$ and related HRL have phoneme similarity it converts both $SL$ and related HRL text into international phonetic alphabet (IPA) symbols.

\subsection{Current Methods' Limitations}
This section discusses four major limitations that existing methods suffer from, and quantifies these with  experimental results. 
The results use the LORELEI dataset~\cite{strassel-tracey-2016-lorelei}, a realistic text corpora that consist of news and social media text, all from  outside-Wikipedia (see 
Section~\ref{sec:exp}).
Some tables also include a comparison with the proposed QL based candidate generation method \query~that we describe in Section~\ref{sec:method}.
We use the definition of \emph{gold candidate recall} in \newcite{zhou2020improving}, which is the proportion of $SL$ mentions that have the gold English entity in the candidate list, as the evaluation metric. 

\subsubsection{Shortage of 
Interlanguage Links}
        As illustrated in Figure~\ref{fig:wiki_size_lor} (specific numbers are in Appendix~\ref{sec:appendix_data}) with statistics from the 2019-10-20 wikidump\footnote{\url{https://dumps.wikimedia.org/}} and in Table~\ref{tbl:sl-en_cover} for five randomly picked low-resource languages,
        many LRL Wikipedias only have a few interlanguage links to the English Wikipedia.
        Consequently, only a few Wikipedia title mappings and anchor text mappings are accessible by all four methods, as shown in  Table~\ref{tbl:method2baseline}.
        
        For \ptm, the workflow is: $SL$ mention $\rightarrow$ $SL$ Wikipedia entity $\rightarrow$ English Wikipedia entity, and it breaks if one link is missing.
        For example, ``Nawala" has an English Wikipedia page, but does not have a corresponding Sinhala page. Given its Sinhala mention\footnote{\url{https://foursquare.com/v/nawala--\%E0\%B6\%B1\%E0\%B7\%80\%E0\%B6\%BD--\%E0\%AE\%A8\%E0\%AE\%B5\%E0\%AE\%B2/4eaa947bf7905c39414c250d/photos}}, the interlink is missing and ~\ptm returns 0 probability for ``Nawala".
        
        For \trans, its translation ability is limited by the tokens contained in the Wikipedia title mappings.
        For a $SL$ mention, when none of its tokens ever appeared in the $SL$ Wikipedia titles, it will not have any English translation, and thus generate no candidates. 
        As for \tsl~and \pivot, they will have fewer data pairs to train on and the model performance would suffer. 
        
        \input{tables/sl_en_wiki_cover}
        \input{tables/mention_token_cover}

        \subsubsection{Small LRL Mention Coverage}
        In the LRL setting, few Wikipedia articles lead to fewer Wikipedia anchor text mappings, thus reducing the ability of current methods to cover many $SL$ mentions.
        For instance, the LRL Oromo Wikipedia article for ``Laayibeeriyaa''\footnote{\url{https://om.wikipedia.org/wiki/Laayibeeriyaa}} has much fewer hyperlinks than the English Wikipedia article for ``Liberia''\footnote{\url{https://en.wikipedia.org/wiki/Liberia}}, even though they are linked through an 
        interlanguage link.
        \Cref{fig:wiki_size_lor,fig:hrl_token_cover_bar} show that \emph{gold candidate recall} generally goes up with the increase in the Wikipedia size. 
        
        To evaluate Wikipedia coverage, we propose a global metric called \emph{mention token coverage} that can be computed without the gold data. 
        \emph{Mention token coverage} 
        is the percentage of $SL$ mentions that have at least one token appearing in Wikipedia title mappings or in anchor text mappings. 
        For example, when we consider Somali language, the mention ``Shabeelada hoose" has a token ``hoose" covered in Wikipedia titles, so it is counted in the \emph{mention token coverage}. But``Soomaalieed" is not covered in Wikipedia titles or anchor text mappings, so it is not counted for \emph{mention token coverage}.
         High \emph{mention token coverage} values tend to guarantee better supervision when trained on Wikipedia. Indeed, in Figure~\ref{fig:hrl_token_cover_bar}, we can clearly see that the \emph{mention token coverage} for LRLs is much smaller than that of high-resource languages (consult Figure~\ref{fig:wiki_size_lor} for the distinction between LRLs and HRLs). 
        
        We also compare the \emph{mention token coverage} with \emph{gold candidate recall} for each method in Figure~\ref{fig:hrl_token_cover_bar}.
        When a method has a \emph{gold candidate recall} higher than \emph{mention token coverage}, it means that the method is able to generalize beyond Wikipedia.
        In contrast, when \emph{gold candidate recall} is lower than \emph{mention token coverage}, it implies that the method is limited by Wikipedia resources.
        
        To compare relation between \emph{mention token coverage} and \emph{gold candidate recall} more clearly, in Figure~\ref{fig:hrl_token_cover_ratio}, we show the ratio of \emph{gold candidate recall} over \emph{mention token coverage}.
        Existing methods' ratio ranges between 0.31 to 1.27, with the average of 0.72. This suggests that existing methods are bounded by Wikipedia resources in most cases. 
        This figure also shows the generalization ability of our QL-based candidate generation method \query~(introduced in Sec.\ref{sec:method}) to outside-Wikipedia mentions with an average ratio of 1.13 ranging between 0.74 and 1.92.
        
        
        \input{tables/translit}
    \subsubsection{
    \tsl~Data Requirements}
        \tsl~suffers from the inability to satisfy several of its data requirements.
        Typically, transliteration models need 
        many training data pairs, which is hard to get 
        using LRL Wikipedia title pairs.
        Also, they require $SL$ mentions and gold English entities to have word-by-word mappings.
        Table \ref{tbl:translit} shows results of \tsl~trained on name pairs from Wikipedia title mappings (\tsl\texttt{-Wiki}).
        We only provide results for 
        languages for which the model has been released; 
        this sample already  shows clearly that the performance of \tsl~drops significantly on LRLs.
        
    \subsubsection{
    \pivot~Prerequisites}
        While \pivot~does not suffer from insufficient cross-lingual resources between $SL$ and English, it is limited by resources between related HRL and English. More importantly, it is limited the availability of related HRL that is similar enough to $SL$. \pivot~learns through grapheme or phoneme similarity. However, grapheme similarity is not enough since not every LRL has a related HRL that uses the same scripts. In these cases \pivot~uses phoneme similarity and maps strings to international phonetic alphabet (IPA) symbols. For example, \cite{zhou2020improving} uses Epitran~\cite{mortensen2018epitran} to convert strings to IPA symbols, but Epitran only supports 55 of the 309 Wikipedia languages, and covers only 8 out of 12 low-resource languages in the LORELEI corpus.
        
        As in Figure~\ref{fig:hrl_token_cover_bar}, \pivot~gains from the language conversion compared with \ptm, but the increase varies among languages, depending on the choice of related HRL. 
        Most importantly, the related HRL cannot replace $SL$ and language conversion may limit on the linking ability. 
        For example, the language pair Oromo-Indonesian(grapheme) has same scripts but \pivot~still suffers from low-resource on Oromo.
        
     

%% file: tables/method_system_table.tex
\begin{table}[h]
\centering
\scriptsize	
{
\setlength{\tabcolsep}{0.6em}
    \begin{tabular}{l|c|c|c|c}
    \toprule[0.7 pt]
    \textbf{Method} & \ptm & \trans & \pivot & \tsl \\
    \hline
    \textbf{Systems} & \multicolumn{4}{c}{Method Usage}\\ 
    \hline
    \chentsemodel & \checkmark & - & - & - \\
    \shyammodel & \checkmark & - & - & - \\
    \hengmodel & - & \checkmark & - & - \\
    \neubmodel & \checkmark & - & \checkmark & - \\
    \hline
    & \multicolumn{4}{c}{Cross-lingual Resource}\\
    \hline
    Title Map & \checkmark & \checkmark & \checkmark & \checkmark \\
    Anchor Text & \checkmark & - & \checkmark & - \\
    Other Lang & - & - & \checkmark & - \\
    \bottomrule[0.7pt]
    \end{tabular}
}
\caption{XEL systems (introduced in Section~\ref{sec:exp}) and Wikipedia cross-lingual resources used by LRL XEL candidate generation methods.}
\vspace{-0.2cm}
\label{tbl:method2baseline}
\end{table}

%% file: tables/sl_en_wiki_cover.tex
\begin{table}[t]
\centering
\small
{
\vspace{-0.3cm}
\setlength{\tabcolsep}{0.3em}
    \begin{tabular}{l|c|c|c|c|c}
    \toprule[0.7pt]
    \textbf{Lang.}   & Oromo  & Akan   & Wolof & Zulu & Somali   \\ \hline
    $SL$-only Entity (\%)  & 34.68  & 70.93  & 37.54 & 49.77 & 12.93  \\    \bottomrule[0.7pt]
    \end{tabular} 
}
\caption{Proportion of gold mentions in the LORELEI corpus with English Wikipedia pages but no $SL$ ones.}
\label{tbl:sl-en_cover}
\end{table}

%% file: tables/mention_token_cover.tex
\begin{figure*}[t]
\centering
\includegraphics[width=1\linewidth]{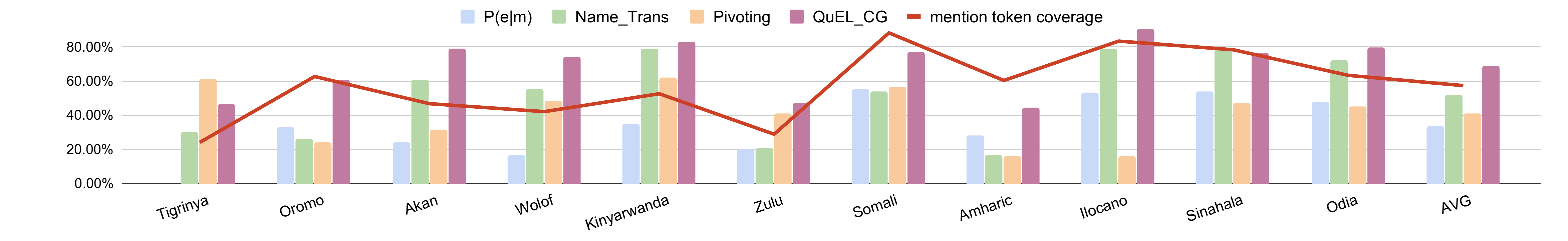}
\includegraphics[width=1\linewidth]{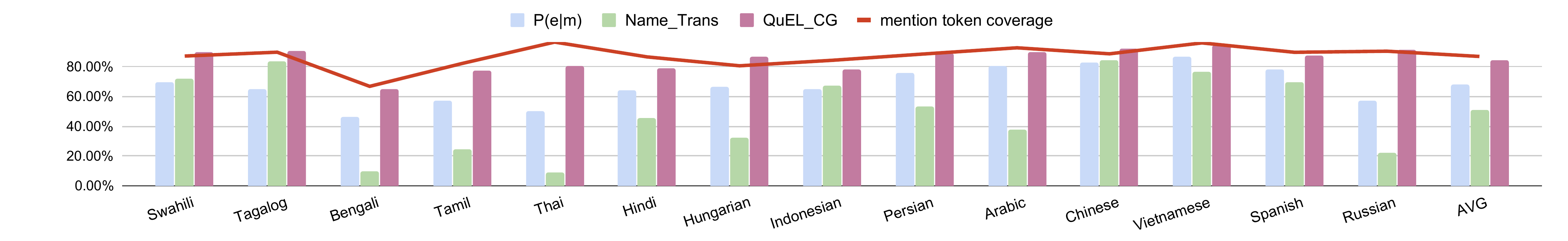}
\vspace{-0.6cm}
\caption{\textbf{{\em Gold candidate recall} (bars) on low-resource languages (top) and high-resource languages (bottom) correlates with {\em mention token coverage} (line) for existing approaches (excluding \query).}
{\em Mention token coverage} is the percentage of $SL$ mentions that have at least one token overlapping with a $SL$ Wikipedia title or anchor text (that correspond to it via interlanguage link).
Languages appear in increasing order of Wikipedia size. 
}
\label{fig:hrl_token_cover_bar}
\end{figure*}


\begin{figure*}[t]
\centering
\includegraphics[width=1\linewidth]{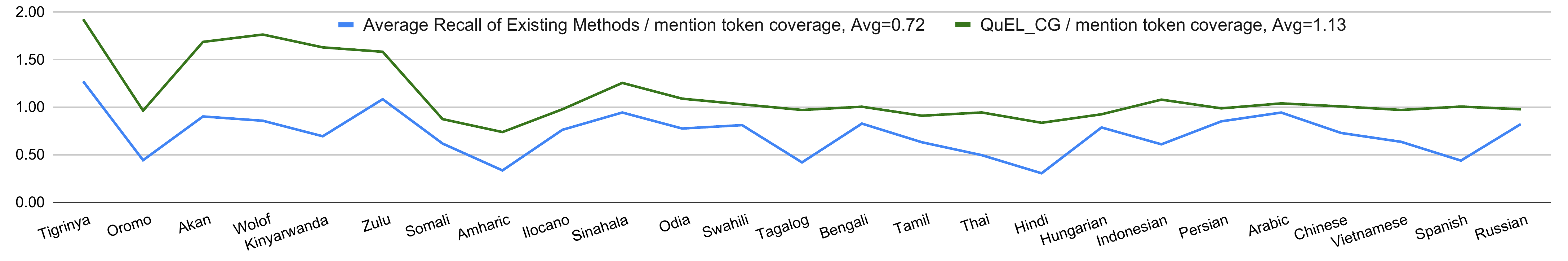}
\vspace{-0.6cm}
\caption{
\textbf{Average {\em gold candidate recall} remains 0.5-0.8 times that of {\em mention token coverage} for existing approaches, but the ratio is 1.0-2.0 for \query.}
The lines are the ratio of {\em gold candidate recall} per candidate generation method divided by {\em mention token coverage}.
Average {\em gold candidate recall} of existing methods: \ptm, \trans, \pivot, is mostly limited by {\em mention token coverage} and cannot exceed it, with 0.72 average ratio. 
However, our proposed \query~can reach recall up to 2 times that of {\em mention token coverage}, with average ratio of 1.13 on all languages, 
tested to be statistically significant with p-value $<$ 0.01\%.}
\label{fig:hrl_token_cover_ratio}
\end{figure*}

%% file: tables/translit.tex
\begin{table}[t]
\centering
\small
{
\vspace{-0.3cm}
\setlength{\tabcolsep}{0.6em}
    \begin{tabular}{l|c|c|c|c}
    \toprule[0.7pt]
    \textbf{Model}  & Hindi & Bengali & Odia & Sinahala  \\ 
    \hline
    \tsl\texttt{-Wiki} & 24.6 & 23.4 & 13.4 & 8.6 \\
    \bottomrule[0.7pt]
    \end{tabular}
}
\caption{{\em Gold candidate recall} of \tsl.}
\label{tbl:translit}
\vspace{-0.2cm}
\end{table}

%% file: sections/4-method.tex
\section{Improved Candidate Generation: \query}
\label{sec:method}
As mentioned in Section~\ref{sec:limit}, Wikipedia's cross-lingual resources are not enough for XEL to perform well in low-resource settings. We argue that outside-Wikipedia resources are essential to compensate for the lack of supervision.
We suggest that search engines query logs provide an excellent resource in this situation, since it is a (very large) super-set of Wikipedia's cross-lingual data but covers even more; as pointed out in the Introduction, it is a collaboratively generated, mildly curated, resource and its effective size is a function of the number of $SL$ native speakers using the search engine.
Thus, query logs can help the candidate generation process map $SL$ mentions to English Wikipedia candidates even when they are not covered by Wikipedia interlanguage links. 

We propose an improved candidate generation method, \query, that uses query log mapping files.
We obtain a high-quality candidate list through directly searching $SL$ mentions in query logs and a query-based pivoting method.
While we can choose any search engines,
we use here Google search\footnote{\url{https://www.google.com/}} and Google maps\footnote{\url{https://www.google.com/maps}}.
\query~ also runs in conjunction with \ptm to cover the cases where query log mapping is not robust.

\subsection{Query Logs as Search Results}
As the first step, we search the morphologically normalized $SL$ mention in the Google search engine
(implementation details in Appendix~\ref{sec:appendix_implementation}) and retrieve a list of web-page results.
We pick top k (k is usually 1 or 5) Wikipedia web-page results $P_k$. 
Note that if the searched result is SL Wikipedia article and it has an interlanguage link to English, we convert it to the corresponding English one through the link, and then mark the corresponding English entity as a candidate. 
When the $SL$ mention is a geopolitical or location entity, we search its normalized mention using Google Map.
Since it only returns English surface of the location instead of Wikipedia articles, we further search the resulted English surface in Google search using the same procedure described above. 

\subsection{Query-based Pivoting}
We also conduct language-indifferent pivoting using query logs. Note that the pivoting methods described below are different from \pivot~in Section~\ref{sec:limit}.

Some LRLs have
high-resource 
languages they are similar to (e.g. Sinhala to Hindi and Tigrinya to Amharic).
In order to exploit the similarities between an LRL and HRL without having to choose one good HRL, we use query logs for pivoting.
We first follow the same search steps described above to get top k Wikipedia web-pages results $P_k$.
Finally, we continue the same process
on the new pivoted mentions.
A special case here is language-specific pivoting on selected language pairs. 
We use a simple utf-8 converter to translate $SL$ mention into a related, but higher-resource language, such as Odia to Hindi, and then run the  candidate generation process described above on the pivoted mention.


%% file: sections/5-exp.tex
\section{Experiments: System Comparison}
\label{sec:exp}
Given the analysis of the key candidate generation process in Section~\ref{sec:limit}, this section moves to study its implications on the overall performance of different LRL XEL systems.
We first propose our LRL XEL framework, \modelname, by combining \query~with a zero-shot candidate ranking module.
Our experimental goal is to compare all systems on both outside-Wikipedia data and Wikipedia data.  
We further analyze the entity distribution and entity type on the linking results.
In addition, an ablation study is demonstrated in Appendix~\ref{sec:appendix_ablation}.

\subsection{Datasets}
Dataset details are reported in Appendix~\ref{sec:appendix_data}.

\noindent    \textbf{LORELEI dataset} \cite{strassel-tracey-2016-lorelei} is a realistic and challenging dataset that includes 
    news and social media such as twitter.
    We divide its 25 languages into LRL and HRL as in Figure~\ref{fig:wiki_size_lor}.
    Entities in LORELEI are of four types: geopolitical entities (GPE), locations (LOC), persons (PER) and organizations (ORG). 
    The dataset provides a specific English KB that mentions are linked to; we processed the original dataset to link to the English Wikipedia instead. Our processed gold labels will be available along with LORELEI dataset\footnote{\url{https://catalog.ldc.upenn.edu/LDC2020T10}}. 
    Given a KB entity, we link it to Wikipedia if the KB provides a Wikipedia link. For a PER or ORG entity without Wikipedia link, we use its KB-provided English information, e.g. name and description, to search for Wikipedia entry, and manually check the correctness. Otherwise, we do not include this entity and remove any mentions linking to it in the EDL dataset. 
    We process these types differently because PER and ORG entities only compose around 5\% of the gold entities.
    
\noindent    \textbf{Wikipedia-based dataset} collected by \cite{TsaiRo16b} is built upon Wikipedia anchor text mappings. 
    All languages in this dataset are high-resource ones as defined in Figure~\ref{fig:wiki_size_lor}.

\subsection{System Comparison}
     We compare the \textbf{supervised} SOTA systems, \chentsemodel, \shyammodel, \hengmodel, \neubmodel, that use candidate generation methods analyzed earlier, with a new, QL-based system, that we present below. Implementation details are in Appendix~\ref{sec:appendix_baseline_implementation}

\subsection{A QL-based XEL: \modelname}
Given the limitations discussed in Sec.~\ref{sec:limit} we propose a new XEL system, \modelname, that uses \query~(Sec.~\ref{sec:method}) along with the following zero-shot candidate ranking module. 
Given a candidate list $C_m$ (the output of \query~on $SL$ mention $m$), \modelname~uses the multilingual BERT \cite{devlin2018bert}~to score the candidates against $m$. 
Specifically, for each candidate $c \in C_m$ it computes a score $W(c, m)$ that measures ``relatedness" between $m$ and $c$.  It then picks the candidate with the highest score as its  output. In case of a tie, we break it by following the candidate selection order, from Google search results, to Google Map results, to \ptm~candidates. We explain below the components of $W(c, m)$.

\noindent\textbf{Candidate Multiplicity Weight}
A candidate can be suggested by multiple sources--Google search, Google Map search, query-based pivoting, or \ptm. \modelname~prefers candidates generated by multiple sources.
We define candidate $c$ multiplicity weight as:
$W_{Source}(c) = Num_{Source}(c)$, the number of sources that generate $c$. 

\noindent\textbf{Contextual Disambiguation}.
\modelname~uses Multilingual BERT (M-BERT) \cite{devlin2018bert} for multilingual embeddings, to compute the similarity of the context of the mention $m$ and the candidate's context in the English Wikipedia.
We denote by $W_{context}(c,m)$ the cosine similarity between $m$'s  context embedding and $c$'s context embedding (see details in Appendix~\ref{sec:appendix_system_implementation}).
Finally, the score for candidate $c$ is $W(c, m) = W_{source}(c) \cdot W_{context}(c,m),$ and  
we select the most likely entity: $e = \argmax_{c \in C_m} W(c,m)$ as output.
\subsection{Entity Linking Results}
\input{tables/performance_figures.tex}

Comprehensive evaluations of both the LORELEI and the Wikipedia based datasets are shown in \Cref{fig:recall_ldc,fig:recall_wiki} and in Table~\ref{tbl:system_dataset_avg}. (Scores that correspond to the figures are reported in Appendix~\ref{sec:appendix_eval}.) 
Note that gold candidate recall on \chentsemodel~and \shyammodel~are identical because they use the same candidate generation module, \ptm.

\modelname~ is shown to significantly improve over existing approaches on both datasets, especially on the more difficult LORELEI dataset where it improves on almost all the languages and shows 
an average of \textbf{25\%} increase in gold candidate recall and of \textbf{13\%} in linking accuracy.
On the Wikipedia-based dataset, \modelname~shows an average increase of \textbf{4\%} in gold candidate recall, while reaching the SOTA on linking accuracy. Importantly, most other systems, (\hengmodel, \chentsemodel~and \shyammodel) use {\em supervised} ranking modules.

%% file: tables/performance_figures.tex
\begin{figure*}[t]
\centering
\includegraphics[width=1\linewidth]{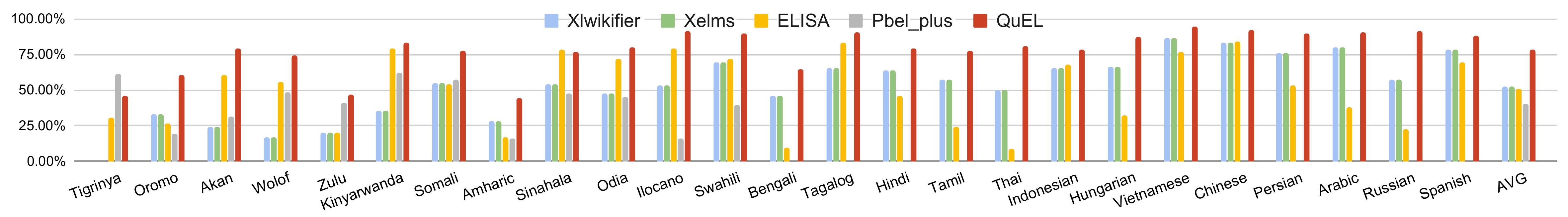}
\includegraphics[width=1\linewidth]{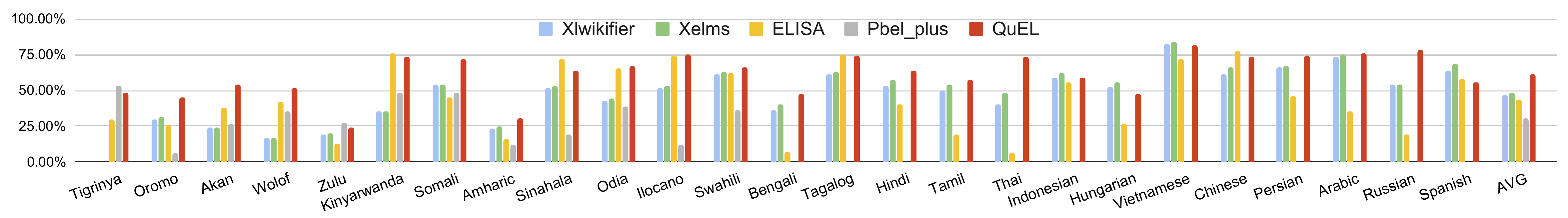}
\vspace{-0.4cm}
\vspace{-0.2cm}
\caption{End-to-end XEL gold candidate recall(top) and linking accuracy(bottom) on the LORELEI dataset sorted by Wikipedia size in ascending order. Specific scores are reported in \Cref{tbl:eval1.1,tbl:eval1.2} in Appendix~\ref{sec:appendix_eval}.}
\label{fig:recall_ldc}
\end{figure*}

\begin{figure*}[t]
\centering
\includegraphics[width=1\linewidth]{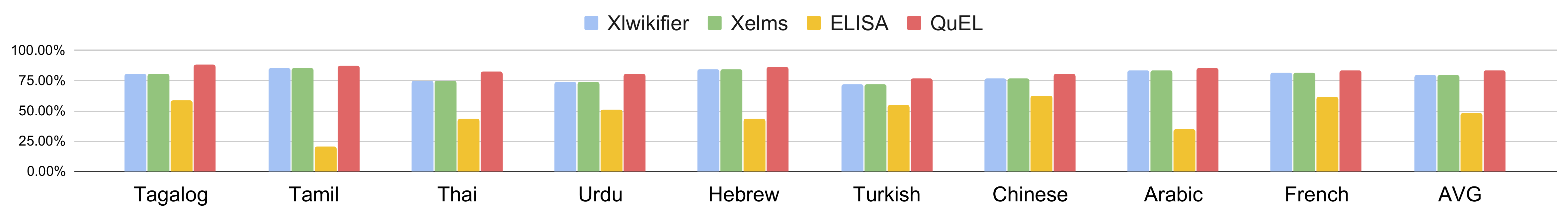}
\includegraphics[width=1\linewidth]{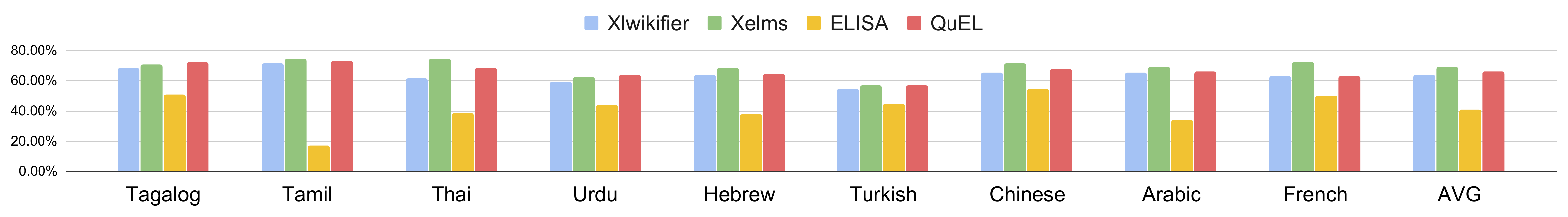}
\vspace{-0.4cm}
\caption{End-to-end XEL gold candidate recall(top) and linking accuracy(bottom) on the Wikipedia-based dataset sorted by Wikipedia size in ascending order. Specific scores are reported in Table~\ref{tbl:eval2} in Appendix~\ref{sec:appendix_eval}.}
\label{fig:recall_wiki}
\end{figure*}

%% file: sections/6-analysis.tex
\begin{table}[h]
\centering
\small
{
\setlength{\tabcolsep}{0.3em}
    \begin{tabular}{l|c|c}
    \toprule[0.7pt]
    \textbf{System}  & Gold Candidate Recall & Linking Accuracy \\ 
    \hline
    & \multicolumn{2}{c}{LORELEI Dataset}\\
    \hline
    \chentsemodel & 52.54 & 46.58\\
    \shyammodel & 52.54 &48.65 \\
    \hengmodel & 50.52 & 43.91\\
    \neubmodel & 38.36 & 30.38\\
    \modelname & \textbf{78.21} & \textbf{61.40}\\
    \hline
    & \multicolumn{2}{c}{Wikipedia-based Dataset}\\
    \hline
    \chentsemodel & 79.40 & 63.51\\
    \shyammodel & 79.40 & \textbf{68.73}\\
    \hengmodel & 47.82 & 41.22\\
    \modelname & \textbf{83.54} & 66.16\\
    \bottomrule[0.7pt]
    \end{tabular}
}
\caption{{\em Gold candidate recall} and {\em linking accuracy} comparison averaged over all languages.
}
\label{tbl:system_dataset_avg}
\end{table}

\input{tables/sl_en_wiki_cover_hit}

\begin{table}[h]
\small
\centering
\vspace{-0.2cm}
\begin{tabular}{cccccc}
\toprule[0.7pt]
\multirow{2}{*}{ \textbf{Lang.} } & \multirow{2}{*}{ \textbf{System} } & \multicolumn{4}{c}{ \textbf{Accuracy (\%)} } \\
& & \textbf{GPE} & \textbf{LOC} & \textbf{PER} & \textbf{ORG} \\
\toprule[0.7pt]
& \shyammodel & 33.0 & 3.6 & 6.2 & 6.3 \\
Oromo& \hengmodel & 37.0 & 11.6 & 12.5 & \textbf{11.9} \\
& \neubmodel & 53.0 & 40.5 & 6.4 & 0.0 \\
 & \modelname & \textbf{57.5} & \textbf{30.4} & \textbf{17.2} & 4.2 \\
\hline
& \shyammodel & 25.6 & 0.0 & 37.5 & 5.7 \\
Zulu& \hengmodel & 11.9 & 2.5 & 31.2 & 5.7 \\
& \neubmodel & 24.6 & 6.2 & 38.1 & 8.7 \\
& \modelname & \textbf{27.8} & \textbf{10.8} & \textbf{50.0} & \textbf{12.6} \\
\hline
& \shyammodel & 55.0 & 28.6 & 21.4 & 42.1 \\
Somali& \neubmodel & 44.7 & 14.3 & \textbf{25.6} & \textbf{75.0} \\
& \hengmodel & 50.1 & 0.0 & 16.7 & 15.0 \\
 & \modelname & \textbf{71.7} & \textbf{57.1} & \textbf{33.3} & 65.0 \\
\bottomrule[0.7pt]
\end{tabular}
\caption{\emph{Linking accuracy} on different types of entities.}
\label{tbl:entity_type}
\end{table}
\vspace{-0.5cm}
\section{Analysis}
Table~\ref{tbl:system_dataset_avg} shows huge performance gaps between the two datasets using the SOTA baseline \shyammodel.
The 20\% percentage difference in both metrics proves that the LORELEI dataset is more difficult due to having more outside-Wikipedia mentions, and more focus on low-resource languages.
One exception is \hengmodel, which has lower performance on Wikipedia-based dataset. We believe it fails to cover many Wikipedia mentions because it does not use Wikipedia anchor text mappings.

Similarly, when we consider the same language performance on the two datasets (see \Cref{fig:recall_ldc,fig:recall_wiki}), e.g., Tamil and Thai, the LORELEI dataset appears harder to deal with.
However, \modelname~achieves similar results on both datasets, and also brings the gold candidate recall for LRL much closer to that of HRL.
Another important observation is that \modelname~performs significantly better on the LORELEI dataset. It suggests that our proposed \query~addresses the outside-Wikipedia coverage problem well by exploiting the query logs.
To understand why \modelname~exceeds baselines largely on LORELEI dataset, we analyze it on entity resource and type distribution as below.

\noindent\textbf{Entity Resource}.
Considering the insufficient Wikipedia interlanguage links for LRL in Table~\ref{tbl:sl-en_cover}, we investigate whether \query~helps in this situation.
In Table~\ref{tbl:sl-en_cover_hit}, \modelname~shows 6.3\% to 27.4\% improvement in gold candidate recall, indicating that it
can effectively perform XEL without the Wikipedia cross-lingual resource, leading to a significant improvement for LRL XEL. 

\noindent\textbf{Entity Type}.
Table~\ref{tbl:entity_type} shows the evaluation on all four types of entities.
We observe that \modelname~improves more on GPE and LOC entities, than on PER and ORG entities. 
We believe the improvement is brought by Google Map query logs. 


%% file: tables/sl_en_wiki_cover_hit.tex
\begin{table}[t]
\centering
\small
{
\vspace{-0.2cm}
\setlength{\tabcolsep}{0.6em}
    \begin{tabular}{l|c|c|c|c|c}
    \toprule[0.7pt]
    \textbf{System} & Akan & Oromo & Zulu & Wolof & Somali \\ 
    \hline
    & \multicolumn{5}{c}{Gold Candidate Recall}\\
    \hline
    \shyammodel & 23.9& 33.2 & 19.8& 16.9 & 55.1 \\
    \hengmodel & 60.7& 26.2& 20.4 & 55.5& 54.3\\
    \neubmodel  & 31.7 & 19.1  & 40.8 &48.5  & 57\\
    \modelname  & \textbf{79.0} & \textbf{60.6} & \textbf{47.1} &  \textbf{74.4} & \textbf{77.4} \\
    \hline
    & \multicolumn{5}{c}{Linking Accuracy}\\
    \hline
    \shyammodel & 23.9& 31.4& 19.8&16.6& 54.5\\
    \hengmodel & 38.00& 25.6 & 12.4&42.2& 45.0\\
    \neubmodel  & 26.5 & 5.9 & 27.7 &35.5& 48.6  \\
    \modelname & \textbf{53.8} & \textbf{45.4} & \textbf{23.8} &\textbf{51.8} & \textbf{72.1}\\
    \bottomrule[0.7pt]
    \end{tabular}
}
\caption{{\em Gold candidate recall} and {\em linking accuracy} for $SL$-only entities in correspondence to Table~\ref{tbl:sl-en_cover}.
}
\label{tbl:sl-en_cover_hit}
\end{table}

%% file: sections/7-conclud.tex
\section{Conclusion}
This work provides a thorough analysis of existing LRL XEL techniques, focusing on the step of generating English candidates for foreign language mentions.
The analysis identifies the inherent lack of sufficient inter-lingual supervision signals as a key shortcoming of the current approaches.
This leads to proposing a new, simple, method that leverages query logs, that are highly effective in addressing these challenges. 
Given that our experiments show a 25\%
increase
in candidate generation,
one future research direction is to
improve candidate ranking in LRL by incorporating coherence statistics and entity types. Moreover, given the effectiveness of query logs, we believe it can be applied to other cross-lingual tasks like relation extraction and Knowledge Base completion.

%% file: sections/7.5-ack.tex
\section*{Acknowledgments}

This work was supported by contracts HR0011-15-C-0113, HR0011-18-2-0052, and FA8750-19-2-1004 with the US Defense Advanced Research Projects Agency (DARPA). The views expressed are those of the authors and do not reflect the official policy or position of the Department of Defense or the U.S. Government.

%% file: sections/8-appendix.tex
\section{Appendices}

\subsection{Dataset Statistics}
\label{sec:appendix_data}
We show the dataset statistics calculated from 2019-10-20 Wikidump\footnote{https://dumps.wikimedia.org} as below, demonstrating the Wikipedia size (article number), interlanguage link size (between $SL$ and English), and test data mention number for every language, on both LORELEI dataset and Wikipedia-based dataset.
\input{tables/dataset_stat.tex}
\newpage

\subsection{Implementation Details of \query}
\label{sec:appendix_implementation}
    To perform our improved candidate generation method \query, we first conduct morphological normalization on the $SL$ mention before querying.
    Then, we use the normalization output as search input.
    We use Google search~\footnote{\url{https://developers.google.com/custom-search/v1/overview}} and Google Map~\footnote{\url{https://cloud.google.com/maps-platform}}.
    We can also customize the search input for better results on extremely low-resource languages (Odia and Illocano) as below.
    
\noindent    \textbf{Morphological normalization}.
    Language-specific Morphological normalization is a basic process for all candidate generation methods.
    An entity may have different surface forms in the document, which makes candidate generation difficult. 
    To cope with this issue, several operations including removing, adding, or replacing suffixes and prefixes are conducted as a prior process. 
    
\noindent    \textbf{Customize search input}. 
    To better retrieve Wikipedia pages as search results and ignore other web-page results, ``wiki" or ``[Country of $SL$]" can be appended to the original search input.

\subsection{Ablation Study}
\label{sec:appendix_ablation}
\input{tables/ablation.tex}

We now quantify the effects of each component in our candidate generation method and show the results in Table~\ref{tbl:ablation}.

\noindent\textbf{Google Map}. Our model is default to use the Google Map cross lingual resource. We test the effect of adding supervision from this QL.

\noindent\textbf{Google top1}. In everywhere that takes the Google query log (QL) results, take only the first Wikipedia page result that is in source or target language as candidate.

\noindent\textbf{Google top5}. Similar to Google top1, take the top 5 Wikipedia page results as candidates. We can see that Google top1 and top5's effects are language dependent.

\noindent\textbf{\ptm}. We test whether adding the \ptm~module would help in linking performance. To better show the results, \ptm~is added under the setting of using QL and Google Map KB, without adding other modules.

\noindent\textbf{Pivoting}. Pivoting here refers to our query-based pivoting, different from \pivot~in Section~\ref{sec:limit}. 
We picked two low-resource languages: Odia and Tigrinya, to explore the pivoting effect and show results in Table~\ref{tbl:ablation_pivot}. 
On Odia, language-specific pivoting skill is used. Since we know in prior that Odia and Hindi are similar while the latter has much more resource, a simple utf8-converter is used to transform Odia to Hindi, and then runs the Hindi mention through our whole system. 
On Tigrinya, a language-indifferent pivoting skill is used.
After getting QL results, besides using Google top1 or top5, we further pick Wikipedia page results that are in any other language, such as Amharic or Scots that have similar scripts, but with richer cross-lingual supervision then Tigrinya.
    
    We further examined the effect of transliteration models using trained models~\cite{UpKoRo18} specifically on Sinhala and Odia, with bilingually mapped Wikipedia titles as supervision. We also used Google transliteration resource for Odia mentions. 
    However, no increase on linking accuracy is observed, and the absolute increase in gold candidate recall is less than 0.5\%.
    Since we only studied on Sinhala and Odia, maybe the transliteration resource is useful on other languages.
    
    \Cref{tbl:ablation,tbl:ablation_pivot} show that we added a lot of value beyond the use of Google search -- simply using google search without adding other parts of our candidate generation methods does not have good linking results. Indeed, incorporating online search engine query logs effectively to XEL is highly non-trivial. In this context, it is important to note that all existing methods make heavy use of Wikipedia, and therefore using QL as a cross-lingual resource is as fair. Moreover, as our results show, the use of Wikipedia allows existing systems to perform well only on Wikipedia data, which is uninteresting for all practical purposes. As shown in Figure~\ref{fig:recall_ldc}, \Cref{tbl:eval1.1,tbl:eval1.2}, our method works well outside Wikipedia!

\begin{table}[H]
\renewcommand{\arraystretch}{0.35}
    \centering
    \small
    \vspace{-0.2cm}
    \setlength{\tabcolsep}{0.3em}
    \begin{tabular}{lccccccccc}
    \toprule[1pt]
\multirow{2}{*}{\textbf{}} & \multicolumn{2}{c}{\textbf{Odia (\%)}}   & \multicolumn{2}{c}{\textbf{Tigrinya (\%)}} \\ \cmidrule[1pt]{2-5} 
\multicolumn{1}{c}{}& \textbf{Accuracy} &  \textbf{Recall} & \textbf{Accuracy} & \textbf{Recall} \\ \midrule[1pt]
        \modelname~w/o pivot.  & 66 & 78.6 & 45.3 & \textbf{46.4}\\
        \midrule
        \modelname  & \textbf{66.7} & \textbf{79.3} & \textbf{45.7} & \textbf{46.4}\\
        \bottomrule
    \end{tabular}
    \caption{Ablation study on 2 low-resource languages to examine effect of pivoting techniques.
    To better show the difference, a simple setting, Google top1 + \ptm, is used for each language.
    }\label{tbl:ablation_pivot}
\end{table}

\subsection{Implementation Details of Compared Systems}
\label{sec:appendix_baseline_implementation}

    For \textbf{\chentsemodel}\footnote{\url{https://github.com/cttsai/illinois-cross-lingual-wikifier}}, we use different versions of candidate ranking on the two datasets.
    Since the Wikipedia-based dataset provides training data, we use its provided version of  candidate ranking.
    However, the LORELEI dataset has no training data, and thus no candidate ranking model can be trained. We just pick the first candidate as the result.
    For a comparison purpose, since \shyammodel~uses the same candidate generation module and provides better candidate ranking, \chentsemodel~is close to and mostly up-bounded by \shyammodel~results.
    
    For \textbf{\shyammodel}\footnote{\url{https://github.com/shyamupa/xelms}}, we use trained ranking modules on most languages when available, and use the zero-shot version of ranking module for the rest of languages (Akan and Kinyarwanda).
    
    For \textbf{\hengmodel}, we access the system using the API~\footnote{\url{https://nlp.cs.rpi.edu/software}} directly provided by its authors, and call the GET/entity\_linking/\{identifier\} function.
    
    For \textbf{\neubmodel}\footnote{\url{https://github.com/shuyanzhou/pbel_plus}}, we test this approach only on low-resource languages on the LORELEI dataset, because it generates candidates through pivoting to a related high-resource language, and it does not make sense to pivot a already high-resource language to other languages.

\subsection{Implementation Details of Our System}
\label{sec:appendix_system_implementation}
During candidate ranking, for a mention $m$ in a document $D$, we get the sentence $s_m$ where $m$ appears and computes its contextualized embedding $v_m = \mbert(e, s_m)$.
For each $c \in C_m$, we retrieve a list of sentences $S_c = \{s_1, s_2, ..., s_n\}$ that contains the candidate entity $c$ in its corresponding Wikipedia page's summary. 
The contextualized embedding for $c$ is denoted by $v_c$: 
$$v_c ={ \frac{1}{\left|S_c\right|}\sum_{i=1}^{n}{\mbert(c, s_i)}}$$

Note that we picked two representative languages: Odia and Ilocano, for which we have additional LORELEI provided monolingual text, and trained the M-BERT model~\cite{devlin2018bert} using their Wikipedia data along with LORELEI text. We did not use pre-trained M-BERT~\footnote{https://github.com/google-research/bert} on all languages because many low-resource languages are not supported, and for the supported ones the performance increase is much less than that of models trained with LORELEI text plus Wikipedia data. This experiment serves to show the gain one could get from additional supervision and, at the same time, highlights the results we show when M-BERT is not available, which is more realistic.

\subsection{Comprehensive Evaluation}
This section includes comprehensive evaluation on XEL systems.

\label{sec:appendix_eval}
\begin{table}[H]
\renewcommand{\arraystretch}{0.4}
    \centering
    \small
    \vspace{-0.3cm}
    \setlength{\tabcolsep}{0.6em}
    \begin{tabular}{llccc}
    \toprule[1pt]
    \textbf{Language} & \textbf{Method} & \textbf{Accu} & \textbf{Rec@5} & \textbf{Rec@n} \\ \midrule[1pt]

    \multirow{4}{*}{Tamil} 
    & \chentsemodel  & 49.8&57.4 & 57.4 \\
    & \shyammodel & 53.8 & 57.4 & 57.4 \\
    & \hengmodel & 19.6 & 24.1 & 24.4\\
    & \modelname & \textbf{58.2} & \textbf{73.6} & \textbf{76.6} \\ \midrule
    \multirow{4}{*}{Zulu} 
    & \chentsemodel  & 19.6& 19.8& 19.8 \\
    & \shyammodel &19.8 & 19.8& 19.8\\
    & \hengmodel & 12.4& 17.9& 20.4 \\
    & \neubmodel & 27.7 & 33.7 &40.8 \\
    & \modelname & \textbf{23.8} & \textbf{41.2} & \textbf{47.1} \\ \midrule
    \multirow{4}{*}{Akan} 
    & \chentsemodel  & 23.9 &23.9& 23.9  \\
    & \shyammodel & 23.9& 23.9& 23.9\\
    & \hengmodel & 38 & 60.5& 60.7 \\ 
    & \neubmodel & 26.5& 28.0 & 31.7 \\
    & \modelname & \textbf{53.8}& \textbf{78.1} &\textbf{79} \\ \midrule
    \multirow{4}{*}{Amharic} 
    & \chentsemodel  & 23.3 &28.2&28.2  \\
    & \shyammodel &24.6 &28.2 &28.2 \\
    & \hengmodel & 16.4&16.7& 16.8 \\  
    & \neubmodel & 11.7& 11.9 & 16.0 \\
    & \modelname &\textbf{30.7} & \textbf{43.8} &\textbf{44.7} \\ \midrule
    \multirow{4}{*}{Hindi} 
    & \chentsemodel  & 53.5& 63.9& 63.9 \\
    & \shyammodel & 57.4& 63.9& 63.9\\
    & \hengmodel & 40.3& 43.4& 45.8 \\        
    & \modelname & \textbf{63.6}& \textbf{74.4} & \textbf{79}\\ \midrule
    \multirow{4}{*}{Indonesian} 
    & \chentsemodel  & 59.2 & 65.3& 65.3 \\
    & \shyammodel & \textbf{62.2}& 65.3& 65.3\\
    & \hengmodel & 56&64& 67.7\\     
    & \modelname &60 & \textbf{73.2} &\textbf{74.6} \\ \midrule
    \multirow{4}{*}{Spanish} 
    & \chentsemodel  &63.9 &78.1&78.1  \\
    & \shyammodel &\textbf{68.4} & 78.1& 78.1\\
    & \hengmodel & 57.8&68.3&69.8 \\
    & \modelname &56 &  \textbf{81.5}& \textbf{87.9}\\ \midrule
    \multirow{4}{*}{Arabic} 
    & \chentsemodel  & 73.3&80.4& 80.4 \\
    & \shyammodel & 75.1& 80.4& 80.4\\
    & \hengmodel & 35.5&37.3&37.9 \\
    & \modelname & \textbf{75.6}& \textbf{84} & \textbf{90.2}\\ \midrule
    \multirow{4}{*}{Swahili} 
    & \chentsemodel  & 61.3&69.6& 69.9 \\
    & \shyammodel &63.4 &69.6 &69.6 \\
    & \hengmodel & 62&71.4&72.2 \\
    & \neubmodel & 36.2 & 37.3 & 39.3 \\
    & \modelname & \textbf{66.3} & \textbf{76.2} & \textbf{76.2}\\   \midrule
    \multirow{4}{*}{Wolof} 
    & \chentsemodel  & 16.6&16.9&16.9  \\
    & \shyammodel &16.6 & 16.9& 16.9\\
    & \hengmodel &42.2 &52.2& 55.5\\
    & \neubmodel & 35.5 & 42.2 & 48.5 \\
    & \modelname & \textbf{51.8} & \textbf{66.1} & \textbf{66.1}\\ \midrule
    \multirow{4}{*}{Vietnamese} 
    & \chentsemodel  &\textbf{82.4} &86.9& 86.9 \\
    & \shyammodel &84.1 & 86.9& 86.9\\
    & \hengmodel &72.1 &76.7& 76.9\\      
    & \modelname & 81.3& \textbf{91.3} & \textbf{95}\\ \midrule
    \multirow{4}{*}{Thai} 
    & \chentsemodel  & 40&50.1&50.1  \\
    & \shyammodel &48.3 &50.1 &50.1 \\
    & \hengmodel & 6.2&9.1&9.1 \\     
    & \modelname &\textbf{73.8} & \textbf{79.4} &\textbf{79.5} \\  \midrule
    \multirow{4}{*}{Bengali} 
    & \chentsemodel  & 36.5 &46.4&46.4  \\
    & \shyammodel & 40.7& 46.4& 46.4\\
    & \hengmodel & 7.3&9.4&9.9 \\  
    & \modelname & \textbf{47.4}& \textbf{61.6} &\textbf{65} \\ \midrule
    \multirow{4}{*}{Tagalog} 
    & \chentsemodel  & 61.4&65.3& 65.3 \\
    & \shyammodel & 63.2& 65.3& 65.3\\
    & \hengmodel & \textbf{75.3}&82.3&83.6 \\  
    & \modelname & 74.1& \textbf{88.5} &\textbf{90.4} \\ \midrule
    \multirow{4}{*}{Hungarian} 
    & \chentsemodel  &52.5 &66.4& 66.4 \\
    & \shyammodel & \textbf{55.8}& 66.4& 66.4\\
    & \hengmodel & 26.3&31.6&32.2 \\
    & \modelname & 47.7& \textbf{78.1} & \textbf{87.2} \\
    \bottomrule
    \end{tabular}
    \caption{Quantitative evaluation results on 25 languages on LORELEI dataset. Accu is linking accuracy, Rec@n is gold candidate recall, with n ranging between 2 to 9 for \modelname~and 100 for \neubmodel. Rec@5 is gold candidate recall if we reserve only top5 candidates by the ranking score.}\label{tbl:eval1.1}
\end{table}

\begin{table}[H]
\renewcommand{\arraystretch}{0.4}
    \centering
    \small
    \vspace{-0.3cm}
    \setlength{\tabcolsep}{0.4em}
    \begin{tabular}{llccc}
    \toprule[1pt]
    \textbf{Language} & \textbf{Method} & \textbf{Accu} & \textbf{Rec@5} & \textbf{Rec@n} \\ \midrule[1pt]

    \multirow{4}{*}{Chinese} 
    & \chentsemodel  & 61.4 & 83.2 & 83.2 \\
    & \shyammodel &66.4 &83.2 &83.2 \\
    & \hengmodel & \textbf{77.3}&83.6&84.5 \\  
    & \modelname & 73.8& \textbf{89.8} &\textbf{92.4} \\ \midrule
    \multirow{4}{*}{Persian} 
    & \chentsemodel  & 66.1 & 76.1 & 76.1  \\
    & \shyammodel & 67& 76.1& 76.1\\
    & \hengmodel & 46.1&53&53.4 \\
    & \modelname & \textbf{74.7}&  \textbf{84.6}& \textbf{89.5}\\ \midrule
    \multirow{4}{*}{Russian} 
    & \chentsemodel  & 53.9 & 57.4 & 57.4 \\
    & \shyammodel & 54.1& 57.4& 57.4\\
    & \hengmodel & 19.1&20.8&22.2 \\  
    & \modelname & \textbf{78.6} &\textbf{87.7}  &\textbf{91.2} \\ \midrule
    \multirow{4}{*}{Oromo} 
    & \chentsemodel  & 29.7 & 33.2 & 33.2 \\
    & \shyammodel & 31.4& 33.2& 33.2\\
    & \hengmodel & 25.6&26.1&26.2 \\   
    & \neubmodel &5.9&20.6&24.2\\    
    & \modelname &\textbf{45.4} & \textbf{57.2} &\textbf{57.2} \\ \midrule
    \multirow{4}{*}{Tigrinya} 
    & \chentsemodel  &0 &0&0  \\
    & \shyammodel & 0& 0&0 \\
    & \hengmodel & 30& 30.4&37 \\
    & \neubmodel &53.4&56.7&61.6\\
    & \modelname &\textbf{45.7} & \textbf{46.4} & \textbf{46.4}\\ \midrule
    \multirow{4}{*}{Sinhala} 
    & \chentsemodel  & 51.9 & 54.1 & 54.1 \\
    & \shyammodel & 52.9& 54.1& 54.1\\
    & \hengmodel &\textbf{72} &\textbf{77.7}& \textbf{78.2}\\ 
      & \neubmodel &19.2&26.7&47.3\\
    & \modelname &64.1 &72.8&76.8 \\ \midrule
    \multirow{4}{*}{Kinyarwanda} 
    & \chentsemodel  & 35.1 & 35.1 & 35.1 \\
    & \shyammodel & 35.1& 35.1& 35.1\\
    & \hengmodel & \textbf{75.9}&79.2&79.2 \\ 
    & \neubmodel & 48.5 & 51.4 & 62.0 \\
    & \modelname & 73.6&  \textbf{83.4}& \textbf{83.4}\\ \midrule
    \multirow{4}{*}{Ilocano} 
    & \chentsemodel  & 52.0 & 53.2 & 53.2 \\
    & \shyammodel & 53.2 & 53.2 & 53.2\\
    & \hengmodel &74.2 &77.4&79.5 \\    
    & \neubmodel &12.3&13.3&16.1\\ 
    & \modelname & \textbf{74.9}& \textbf{84.9} & \textbf{91.1}\\ \midrule
    \multirow{4}{*}{Odia} 
    & \chentsemodel  & 42.6&47.6&47.6  \\
    & \shyammodel & 44.29& 47.6&47.6 \\
    & \hengmodel & 65.1&71.8&72.3 \\
    & \neubmodel &39.1&42.0&45.5\\ 
    & \modelname &\textbf{66.7} & \textbf{79.2} & \textbf{79.7}\\ \midrule
    \multirow{4}{*}{Somali} 
    & \chentsemodel  &54.5 &55.1& 55.1 \\
    & \shyammodel & 54.5& 55.1&55.1 \\
    & \hengmodel & 45&53.1&54.3 \\        
    & \neubmodel & 48.6 & 54.5 & 57.0 \\
    & \modelname & \textbf{71.2}& \textbf{80.7} & \textbf{81.5}\\
    \bottomrule
    \end{tabular}
    \caption{Quantitative evaluation results on 25 languages on LORELEI dataset (continued). Accu is linking accuracy, Rec@n is gold candidate recall, with n ranging between 2 to 9 for \modelname~and 100 for \neubmodel. Rec@5 is gold candidate recall if we reserve only top5 candidates by the ranking score.}
    \label{tbl:eval1.2}
\end{table}

\newpage
\begin{table}[H]
\renewcommand{\arraystretch}{0.4}
    \centering
    \small
    \vspace{-0.3cm}
    \setlength{\tabcolsep}{0.4em}
    \begin{tabular}{llccc}
    \toprule[1pt]
    \textbf{Language} & \textbf{Method} & \textbf{Accu} & \textbf{Rec@5} & \textbf{Rec@n} \\ \midrule[1pt]

    \multirow{4}{*}{Arabic\_wiki} 
    & \chentsemodel  & 65.2&83.4&83.4  \\
    & \shyammodel & \textbf{69.2}&83.4&83.4\\
    & \hengmodel & 34.1&34.5&34.9 \\
    & \modelname & 66.1& \textbf{85.2} & \textbf{85.5}\\ \midrule
    \multirow{4}{*}{French\_wiki} 
    & \chentsemodel  &62.7  & 81.1 & 81.9 \\
    & \shyammodel & \textbf{71.8}& 81.6& 81.9\\
    & \hengmodel & 50.3&59&61.2 \\     
    & \modelname & 63.2 & \textbf{82.5} & \textbf{83.5} \\ \midrule
    \multirow{4}{*}{Hebrew\_wiki} 
    & \chentsemodel  & 63.5&84.4&84.9  \\
    & \shyammodel & \textbf{68.4}& 84.9& 84.39\\
    & \hengmodel & 37.39&42.4& 43.2\\  
    & \modelname & 64.6& \textbf{86.3} & \textbf{86.8}\\ \midrule
    \multirow{4}{*}{Tamil\_wiki} 
    & \chentsemodel  & 71.5&85.6&85.8  \\
    & \shyammodel &\textbf{74.1} &85.8&85.8 \\
    & \hengmodel & 16.9&20&20.5 \\  
    & \modelname & 72.8 & \textbf{87.4}& \textbf{87.5} \\ \midrule
    \multirow{4}{*}{Thai\_wiki} 
    & \chentsemodel  &73.36 &75.1&75.3  \\
    & \shyammodel & \textbf{74.5}& 75.3& 75.3\\
    & \hengmodel & 38.9&42.6&43.5 \\ 
    & \modelname & 68.1& \textbf{82.4} & \textbf{82.4}\\ \midrule
    \multirow{4}{*}{Tagalog\_wiki} 
    & \chentsemodel  & 68&80.3&80.4  \\
    & \shyammodel & 70.5&80.4 &80.4 \\
    & \hengmodel & 50.5&57.5&58.8 \\  
    & \modelname & \textbf{72.3}& \textbf{88.6} &\textbf{88.7} \\ \midrule
    \multirow{4}{*}{Turkish\_wiki} 
    & \chentsemodel  & 54.5&72.1&72.5  \\
    & \shyammodel & 56.8&72.4 &72.5 \\
    & \hengmodel & 44.5&52.2&54.7 \\
    & \modelname & \textbf{57.1} & \textbf{76.1}& \textbf{76.5}\\ \midrule
    \multirow{4}{*}{Urdu\_wiki} 
    & \chentsemodel  & 59.5&73.2&73.5  \\
    & \shyammodel & 62.4& 73.5& 73.5\\
    & \hengmodel & 43.6&50.1&51 \\
    & \modelname &\textbf{63.8} & \textbf{80.4} & \textbf{80.7}\\ \midrule
    \multirow{4}{*}{Chinese\_wiki} 
    & \chentsemodel  & 64.9&76.8&76.9  \\
    & \shyammodel & \textbf{71.2} & 76.9& 76.9\\
    & \hengmodel & 54.3&60.4&62.6 \\
    & \modelname & 67.4& \textbf{80.3} & \textbf{80.3}\\
    \bottomrule
    \end{tabular}
    \caption{Quantitative evaluation results on 9 languages on Wikipedia-based dataset. Accu is linking accuracy, Rec@n is gold candidate recall, with n ranging between 2 to 9 for \modelname~and 100 for \neubmodel. Rec@5 is gold candidate recall if we reserve only top5 candidates by the ranking score.}\label{tbl:eval2}
\end{table}

%% file: tables/dataset_stat.tex
\begin{table}[H]
    \centering
    \small
    {
    \vspace{-0.3cm}
    \setlength{\tabcolsep}{0.5em}
    \begin{tabular}{l c c}
    \toprule[1pt]
    {\bf Language}&{\bf Interlang./Wiki}&{\bf \#Test Mentions}\\
    \midrule[1pt]
    Tigrinya& 189/226 & 3174  \\ \midrule
    Oromo& 621/790 & 2576  \\ \midrule
    Akan& 726/815 & 462  \\ \midrule
    Wolof& 1,231/1389 & 302  \\ \midrule
    Zulu& 1,328/2221 & 1071  \\ \midrule
    Kinyarwanda& 1,670/1826 & 521  \\ \midrule
    Somali& 4,025/5,831 & 884  \\ \midrule
    Amharic& 8,176/14,856 & 1157  \\ \midrule
    Sinahala& 11,314/15,722 & 673  \\ \midrule
    Odia& 12,307/15,767 & 2079  \\ \midrule
    Ilocano& 12,377/15,200 & 1274  \\ \midrule
    Swahili& 34,354/59,040 & 1251  \\ \midrule
    Bengali& 64,183/89,095 & 1266  \\ \midrule
    Tagalog& 64,847/72,103 & 1050  \\ \midrule
    Hindi& 74,906/139,328 & 814  \\ \midrule
    Tamil& 76,800/129,281 & 1157  \\ \midrule
    Thai& 98,088/138,252 & 1122  \\ \midrule
    Indonesian& 286,723/531,808 & 1376  \\ \midrule
    Hungarian& 331,829/471,164 & 1059  \\ \midrule
    Vietnamese& 550,111/1,246,441 & 990  \\ \midrule
    Chinese& 612,335/1,122,232 & 1157  \\ \midrule
    Persian& 603,740/728,638 & 877  \\ \midrule
    Arabic& 633,168/1,046,282 & 1188  \\ \midrule
    Russian& 847,036/1,630,773 & 1205  \\ \midrule
    Spanish& 1,005,407/1,602,399 & 711  \\
    \bottomrule
\end{tabular}
}
    \caption{Overview for LORELEI dataset. Wiki/Interlang. refers to corresponding Wikipedia and English interlanguage link size.}
    \label{tbl:dataset1}
\end{table}

\begin{table}[H]
    \centering
    \small
    {
    \vspace{-0.3cm}
    \setlength{\tabcolsep}{0.5em}
    \begin{tabular}{l c c}
    \toprule[1pt]
    {\bf Language }&{\bf Interlang./Wiki}&{\bf \# Test Mentions}\\
    \midrule[1pt]
    Tagalog& 64,847/72,103 & 1,075  \\ \midrule
    Tamil& 76,800/129,281 & 1,075  \\ \midrule
    Thai& 98,088/138,252 & 11,380  \\ \midrule
    Urdu& 128,227/154,103 & 1,390  \\ \midrule
    Hebrew& 193,391/267,243 & 16,137  \\ \midrule
    Turkish& 244,882/354,767 & 13,795  \\ \midrule
    Chinese& 612,335/1,122,232 & 11,252  \\ \midrule
    Arabic& 633,168/1,046,282 & 1,0647  \\ \midrule
    French& 1,398,118/2,221,709 & 2,637  \\
    \bottomrule
\end{tabular}
}
    \caption{Overview for Wikipedia-based dataset. Wiki/Interlang. refers to corresponding Wikipedia and English interlanguage link size.}
    \label{tbl:dataset2}
\end{table}

%% file: tables/ablation.tex
\begin{table*}[t]
\renewcommand{\arraystretch}{0.4}
    \centering
    \small
    \vspace{-0.3cm}
    \setlength{\tabcolsep}{0.7em}
    \begin{tabular}{lcccccc}
    \toprule[1pt]
\multirow{2}{*}{\textbf{}} & \multicolumn{2}{c}{\textbf{Akan (\%)}}   & \multicolumn{2}{c}{\textbf{Thai (\%)}} & \multicolumn{2}{c}{\textbf{Tigrinya (\%)}} \\ \cmidrule[1pt]{2-7}
\multicolumn{1}{c}{}& \textbf{Accuracy} &  \textbf{Cand. Recall} & \textbf{Accuracy} & \textbf{Cand. Recall}& \textbf{Accuracy} & \textbf{Cand. Recall} \\ \midrule[1pt]
        Google top1 w/o Google Map & 53.4 & 57.9 & 73.5 & 74.6 & 31.7 & 31.9 \\
        \midrule
        Google top1 & 54.7 & 61 & \textbf{74} & 77.1& \textbf{44.9} & 46.4\\
        \midrule
        Google top5 & 54 & 79 &73.8 & 79.5 & 37.2 & 48.7\\
        \midrule
        Google top1 + \ptm & \textbf{55.1} & 61 & 73.8 & 78.8 & 45.3 & 46.4\\
        \midrule
        Google top5 + \ptm &53.8 & \textbf{79} &73.5& \textbf{80.9} & 36.3 & \textbf{48.7}\\
    \midrule[1pt]
    \toprule
\multirow{2}{*}{\textbf{}} & \multicolumn{2}{c}{\textbf{Oromo (\%)}}   & \multicolumn{2}{c}{\textbf{Somali (\%)}} & \multicolumn{2}{c}{\textbf{Oria (\%)}} \\ \cmidrule[1pt]{2-7} 
\multicolumn{1}{c}{}& \textbf{Accuracy} &  \textbf{Cand. Recall} & \textbf{Accuracy} & \textbf{Cand. Recall}& \textbf{Accuracy} & \textbf{Cand. Recall} \\ \midrule[1pt]
        Google top1 w/o Google Map & 40 & 43.8 & 67.8 & 71.4 & 47.6 & 55.3\\
        \midrule
        Google top1 & 43.9 &50.1  &71.8 &	76.3 & 59.2& 70\\
        \midrule
        Google top5 &41.8 & 55.5 &70.7 & 80.6& 56.5 &76.6 \\
        \midrule
            Google top1 + \ptm & \textbf{45.4} & 57.2 & \textbf{72.1} &	77.4& \textbf{66} & 78.6\\
        \midrule
        Google top5 + \ptm & 42.7 & \textbf{60.6} & 71.2 & \textbf{81.5} & 64.6 & \textbf{83.5}\\
        \bottomrule
    \end{tabular}
    \caption{Ablation study on 6 low-resource languages that examines each candidate generation component for end-to-end linking accuracy (left) and gold candidate recall(right). Candidate number is below 5 in most languages and varies between 2-9. Our method as default includes Google Map module.}\label{tbl:ablation}
\end{table*}

%% file: emnlp2020.bbl
\begin{thebibliography}{18}
\expandafter\ifx\csname natexlab\endcsname\relax\def\natexlab#1{#1}\fi

\bibitem[{Alfonseca et~al.(2010)Alfonseca, Pasca, and
  Robledo-Arnuncio}]{alfonseca2010acquisition}
Enrique Alfonseca, Marius Pasca, and Enrique Robledo-Arnuncio. 2010.
\newblock Acquisition of instance attributes via labeled and related instances.
\newblock In \emph{Proceedings of the 33rd international ACM SIGIR conference
  on Research and development in information retrieval}, pages 58--65.

\bibitem[{Devlin et~al.(2018)Devlin, Chang, Lee, and
  Toutanova}]{devlin2018bert}
Jacob Devlin, Ming-Wei Chang, Kenton Lee, and Kristina Toutanova. 2018.
\newblock Bert: Pre-training of deep bidirectional transformers for language
  understanding.
\newblock \emph{arXiv preprint arXiv:1810.04805}.

\bibitem[{Dredze et~al.(2010)Dredze, McNamee, Rao, Gerber, and
  Finin}]{Dredze:2010:EDK:1873781.1873813}
Mark Dredze, Paul McNamee, Delip Rao, Adam Gerber, and Tim Finin. 2010.
\newblock Entity disambiguation for knowledge base population.
\newblock In \emph{Proceedings of the 23rd International Conference on
  Computational Linguistics}, pages 277--285.

\bibitem[{Monahan et~al.(2011)Monahan, Lehmann, Nyberg, Plymale, and
  Jung}]{Monahan2011CrossLingualCC}
Sean Monahan, John Lehmann, Timothy Nyberg, Jesse Plymale, and Arnold Jung.
  2011.
\newblock Cross-lingual cross-document coreference with entity linking.
\newblock In \emph{TAC}.

\bibitem[{Mortensen et~al.(2018)Mortensen, Dalmia, and
  Littell}]{mortensen2018epitran}
David~R Mortensen, Siddharth Dalmia, and Patrick Littell. 2018.
\newblock Epitran: Precision g2p for many languages.
\newblock In \emph{Proceedings of the Eleventh International Conference on
  Language Resources and Evaluation (LREC 2018)}.

\bibitem[{Pan et~al.(2017)Pan, Zhang, May, Nothman, Knight, and
  Ji}]{pan2017cross}
Xiaoman Pan, Boliang Zhang, Jonathan May, Joel Nothman, Kevin Knight, and Heng
  Ji. 2017.
\newblock Cross-lingual name tagging and linking for 282 languages.
\newblock In \emph{Proceedings of the 55th Annual Meeting of the Association
  for Computational Linguistics (Volume 1: Long Papers)}, volume~1, pages
  1946--1958.

\bibitem[{Pantel et~al.(2012)Pantel, Lin, and Gamon}]{pantel2012mining}
Patrick Pantel, Thomas Lin, and Michael Gamon. 2012.
\newblock Mining entity types from query logs via user intent modeling.
\newblock In \emph{Proceedings of the 50th Annual Meeting of the Association
  for Computational Linguistics: Long Papers-Volume 1}, pages 563--571.
  Association for Computational Linguistics.

\bibitem[{Rijhwani et~al.(2019)Rijhwani, Xie, Neubig, and
  Carbonell}]{rijhwani2019zero}
Shruti Rijhwani, Jiateng Xie, Graham Neubig, and Jaime Carbonell. 2019.
\newblock Zero-shot neural transfer for cross-lingual entity linking.
\newblock In \emph{Proceedings of the AAAI Conference on Artificial
  Intelligence}, volume~33, pages 6924--6931.

\bibitem[{R{\"u}d et~al.(2011)R{\"u}d, Ciaramita, M{\"u}ller, and
  Sch{\"u}tze}]{rud2011piggyback}
Stefan R{\"u}d, Massimiliano Ciaramita, Jens M{\"u}ller, and Hinrich
  Sch{\"u}tze. 2011.
\newblock Piggyback: Using search engines for robust cross-domain named entity
  recognition.
\newblock In \emph{Proceedings of the 49th Annual Meeting of the Association
  for Computational Linguistics: Human Language Technologies-Volume 1}, pages
  965--975. Association for Computational Linguistics.

\bibitem[{{Shen} et~al.(2015){Shen}, {Wang}, and {Han}}]{6823700}
W.~{Shen}, J.~{Wang}, and J.~{Han}. 2015.
\newblock Entity linking with a knowledge base: Issues, techniques, and
  solutions.
\newblock \emph{IEEE Transactions on Knowledge and Data Engineering},
  27(2):443--460.

\bibitem[{Strassel and Tracey(2016)}]{strassel-tracey-2016-lorelei}
Stephanie Strassel and Jennifer Tracey. 2016.
\newblock {LORELEI} language packs: Data, tools, and resources for technology
  development in low resource languages.
\newblock In \emph{Proceedings of the Tenth International Conference on
  Language Resources and Evaluation}, pages 3273--3280.

\bibitem[{Tsai and Roth(2016)}]{TsaiRo16b}
Chen-Tse Tsai and Dan Roth. 2016.
\newblock \href {http://cogcomp.org/papers/TsaiRo16b.pdf} {{Cross-lingual
  Wikification Using Multilingual Embeddings}}.
\newblock In \emph{Proc. of the Annual Conference of the North American Chapter
  of the Association for Computational Linguistics (NAACL)}.

\bibitem[{Tsai and Roth(2018)}]{TsaiRo18}
Chen-Tse Tsai and Dan Roth. 2018.
\newblock \href {http://cogcomp.org/papers/TsaiRo18.pdf} {{Learning Better Name
  Translation for Cross-Lingual Wikification}}.
\newblock In \emph{Proc. of the Conference on Artificial Intelligence (AAAI)}.

\bibitem[{Upadhyay et~al.(2018{\natexlab{a}})Upadhyay, Gupta, and
  Roth}]{UpGuRo18}
Shyam Upadhyay, Nitish Gupta, and Dan Roth. 2018{\natexlab{a}}.
\newblock \href {http://cogcomp.org/papers/UpGuRo18.pdf} {{Joint Multilingual
  Supervision for Cross-Lingual Entity Linking}}.
\newblock In \emph{Proc. of the Conference on Empirical Methods in Natural
  Language Processing (EMNLP)}.

\bibitem[{Upadhyay et~al.(2018{\natexlab{b}})Upadhyay, Kodner, and
  Roth}]{UpKoRo18}
Shyam Upadhyay, Jordan Kodner, and Dan Roth. 2018{\natexlab{b}}.
\newblock \href {http://cogcomp.org/papers/UpKoRo18.pdf} {{Bootstrapping
  Transliteration with Constrained Discovery for Low-Resource Languages}}.
\newblock In \emph{Proc. of the Conference on Empirical Methods in Natural
  Language Processing (EMNLP)}.

\bibitem[{Zhang et~al.(2018)Zhang, Lin, Pan, Lu, May, Knight, and
  Ji}]{zhang2018elisa}
Boliang Zhang, Ying Lin, Xiaoman Pan, Di~Lu, Jonathan May, Kevin Knight, and
  Heng Ji. 2018.
\newblock Elisa-edl: A cross-lingual entity extraction, linking and
  localization system.
\newblock In \emph{Proceedings of the 2018 Conference of the North American
  Chapter of the Association for Computational Linguistics: Demonstrations},
  pages 41--45.

\bibitem[{Zhou et~al.(2020)Zhou, Rijhawani, Wieting, Carbonell, and
  Neubig}]{zhou2020improving}
Shuyan Zhou, Shruti Rijhawani, John Wieting, Jaime Carbonell, and Graham
  Neubig. 2020.
\newblock Improving candidate generation for low-resource cross-lingual entity
  linking.
\newblock \emph{arXiv preprint arXiv:2003.01343}.

\bibitem[{Zhou et~al.(2019)Zhou, Rijhwani, and Neubig}]{zhou2019towards}
Shuyan Zhou, Shruti Rijhwani, and Graham Neubig. 2019.
\newblock Towards zero-resource cross-lingual entity linking.
\newblock \emph{arXiv preprint arXiv:1909.13180}.

\end{thebibliography}
